\documentclass[10pt,twocolumn,letterpaper]{article}

\usepackage{cvpr}              

\usepackage{graphicx}
\usepackage{amsmath}
\usepackage{amssymb}
\usepackage{booktabs}
\usepackage{subcaption}

\usepackage[pagebackref,breaklinks,colorlinks]{hyperref}

\usepackage[capitalize]{cleveref}
\crefname{section}{Sec.}{Secs.}
\Crefname{section}{Section}{Sections}
\Crefname{table}{Table}{Tables}
\crefname{table}{Tab.}{Tabs.}

\def\cvprPaperID{13} 
\def\confName{CVPR}
\def\confYear{2022}

\begin{document}

\title{Semantic Segmentation for Thermal Images: A Comparative Survey}

\author{Zülfiye Kütük \qquad\qquad Görkem Algan\\
Dept. of Image Proc. \& Computer Vis. Technologies, Aselsan Inc., Turkey\\
{\tt\small \{zulfiyekutuk, galgan\}@aselsan.com.tr}
}

\maketitle

\begin{abstract}
Semantic segmentation is a challenging task since it requires excessively more low-level spatial information of the image compared to other computer vision problems. The accuracy of pixel-level classification can be affected by many factors, such as imaging limitations and the ambiguity of object boundaries in an image. Conventional methods exploit three-channel RGB images captured in the visible spectrum with deep neural networks (DNN). Thermal images can significantly contribute during the segmentation since thermal imaging cameras are capable of capturing details despite the weather and illumination conditions. Using infrared spectrum in semantic segmentation has many real-world use cases, such as autonomous driving, medical imaging, agriculture, defense industry, etc. Due to this wide range of use cases, designing accurate semantic segmentation algorithms with the help of infrared spectrum is an important challenge. One approach is to use both visible and infrared spectrum images as inputs. These methods can accomplish higher accuracy due to enriched input information, with the cost of extra effort for the alignment and processing of multiple inputs. Another approach is to use only thermal images, enabling less hardware cost for smaller use cases. Even though there are multiple surveys on semantic segmentation methods, the literature lacks a comprehensive survey centered explicitly around semantic segmentation using infrared spectrum. This work aims to fill this gap by presenting algorithms in the literature and categorizing them by their input images.
\end{abstract}

\section{Introduction}
\label{sec:intro}

Semantic segmentation is one of the high-level tasks in computer vision that assigns a label for each pixel of an image. Semantic segmentation plays a significant role in many applications since it is able to provide scene understanding at the pixel level. Some of those applications include pedestrian segmentation, autonomous driving, and medical diagnosis. Semantic segmentation differs from other common computer vision tasks such as image classification and object detection in terms of its output. For instance, image classification provides which object exists in an image, while object detection gives the object labels and locations by a bounding box. Image segmentation is divided into three sub-branches: semantic, instance, and panoptic segmentation. Semantic segmentation provides a class label for each pixel of an image, while instance segmentation identifies and segments each instance of a class separately. Moreover, panoptic segmentation aims to find the class label for every pixel in an image and all the instances. 

The interest in semantic segmentation has increased rapidly since the deep learning methods achieved promising results. In other words, deep learning-based semantic segmentation approaches have demonstrated a significant boost in efficiency compared to older methods. Different DNN architectures and mechanisms are proposed to obtain better segmentation results. For instance, fully convolutional networks (FCN)\cite{long2015fully} have led to recent advances in deep learning-based semantic segmentation since many novel models use it to get dense predictions. Many semantic segmentation networks also employ the encoder-decoder structure. The encoders extract the features by reducing the resolution, and the decoders restore the resolution. SegNet \cite{badrinarayanan2017segnet} passes the indices of the max locations during pooling in the encoder to the decoder. Also, Unet \cite{ronneberger2015u} employs an encoder-decoder structure with skip connections that pass high-resolution features from the contracting path to the expanding path to guide semantic segmentation. DeepLabV3+ \cite{chen2018encoder} benefits from spatial pyramid pooling and atrous convolution mechanisms. In addition, BiseNetV2 \cite{yu2021bisenet} captures high-level semantics and spatial details with two-pathway architecture. An aggregation layer is also used to exploit these extracted features for the semantic segmentation task.

Most of the methods in the literature exploit three-channel RGB images captured by visible cameras. However, due to the visible imaging limitations, these methods cannot provide the desired performance in adverse environmental conditions such as low-illuminated, rainy, foggy. Therefore, thermal images have been utilized for semantic segmentation tasks since thermal cameras capture thermal radiation, which is more stable at any weather and time. However, thermal images usually have low resolution and ambiguous object boundaries caused by thermal crossover, a phenomenon where the thermal radiation coming from two different objects cannot be distinguished. Moreover, thermal crossover and the lack of the thermal dataset cause semantic segmentation in thermal images to be under-explored. 

To the best of our knowledge, this work is the first survey of semantic segmentation methods in thermal images. The key contributions of this survey are as follows:
\begin{itemize}
    \setlength\itemsep{0em}
    \item A broad survey of current datasets, including RGB and thermal (RGB-T) image pairs and solely thermal images.
    \item A comprehensive review of the deep learning-based thermal image semantic segmentation methods with their architectures and contributions. 
    \item A well-organized comparison of the methods with the announced quantitative measures in the papers.  
\end{itemize}

This paper is organized as follows; Section II overviews the datasets, including thermal and RGB images, deep learning-based semantic segmentation methods for multi-spectral inputs, and a brief discussion on the presented methods. Section III introduces thermal image datasets, semantic segmentation methods using only thermal images, and a comparison of the methods with quantitative measures according to the revealed results in the papers. Finally, Section IV concludes this survey.  

\section{Combining Infrared and Visible Spectrum for Semantic Segmentation}
The fact that infrared and visible spectrum information is in different light spectrums allows them to compensate for each other's deficiencies. While this provides an advantage, it restricts the use cases of the proposed algorithms only on specific hardware with two different sensors for thermal and visible light. Moreover, these methods require additional algorithms to fuse information coming from different spectrums. The fusion methods should avoid information conflicts while incorporating complementary information from different modalities. Besides, few datasets provide RGB-T aligned images with annotations. The number of proposed methods is limited due to the reasons mentioned above.

Utilizing RGB and thermal images simultaneously improves the model's performance. Therefore, this part mentions datasets with RGB-T images and segmentation methods using both visible and infrared spectrums. 

\subsection{Multi-spectral Datasets}
Multi-Spectral Fusion Networks (MFNet) Dataset \cite{ha2017mfnet} contains both RGB and IR images captured using an InfRec R500 camera. This camera has different lenses and sensors for visible and infrared spectrum. The spatial resolutions of all images are 480x640. The dataset consists of 820 daytime and 749 nighttime urban scene images annotated with eight classes (car, person, bike, curve, car stop, guardrail, color cone, and bump). Moreover, the training set contains 50\% of the daytime images and 50\% of the nighttime images, while the remaining images are separated equally for the validation and test sets. Some prediction results of MFNet \cite{ha2017mfnet} and SegNet \cite{badrinarayanan2017segnet} can be seen in Figure \ref{fig:mfnet} which is directly taken from \cite{ha2017mfnet}. Also, RGB-T image pairs and ground truth annotations from the dataset are presented in the first three rows of Figure \ref{fig:mfnet}.

\begin{figure}[h]
    \centering
        \begin{subfigure}{2.5cm}
        \includegraphics[width=2.5cm]{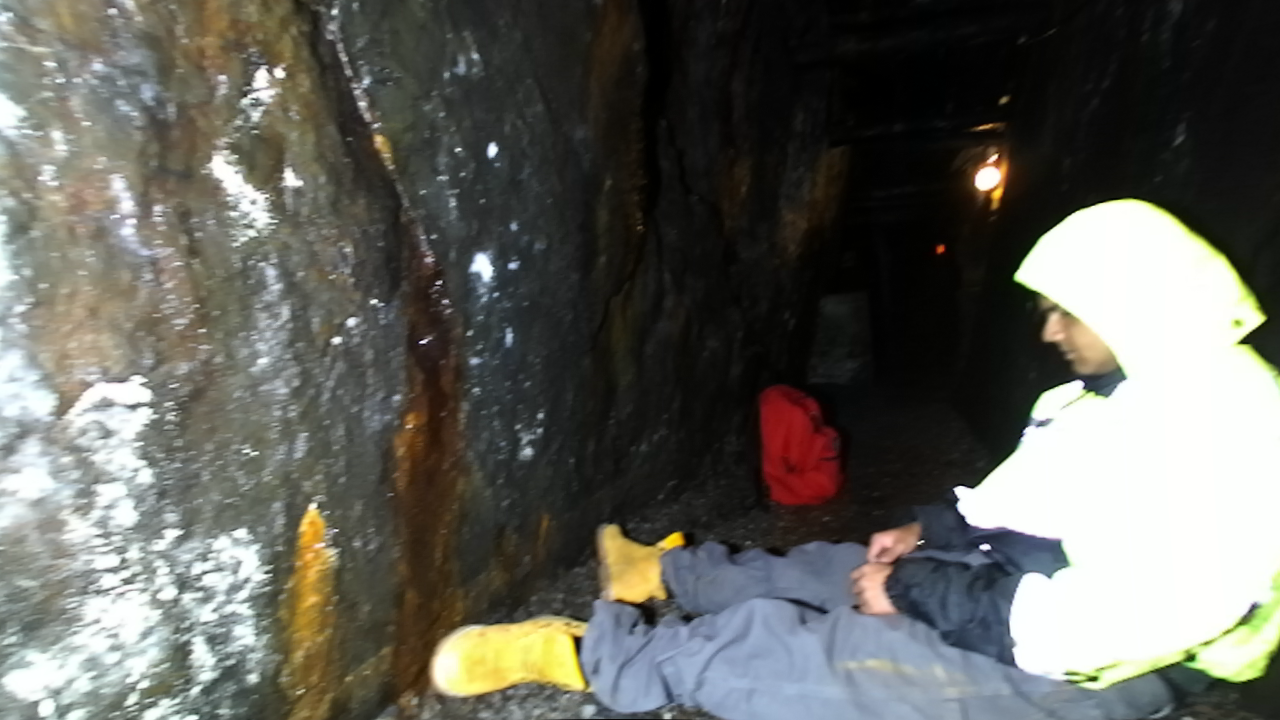} 
        \end{subfigure}
        \begin{subfigure}{2.5cm}
        \includegraphics[width=2.5cm]{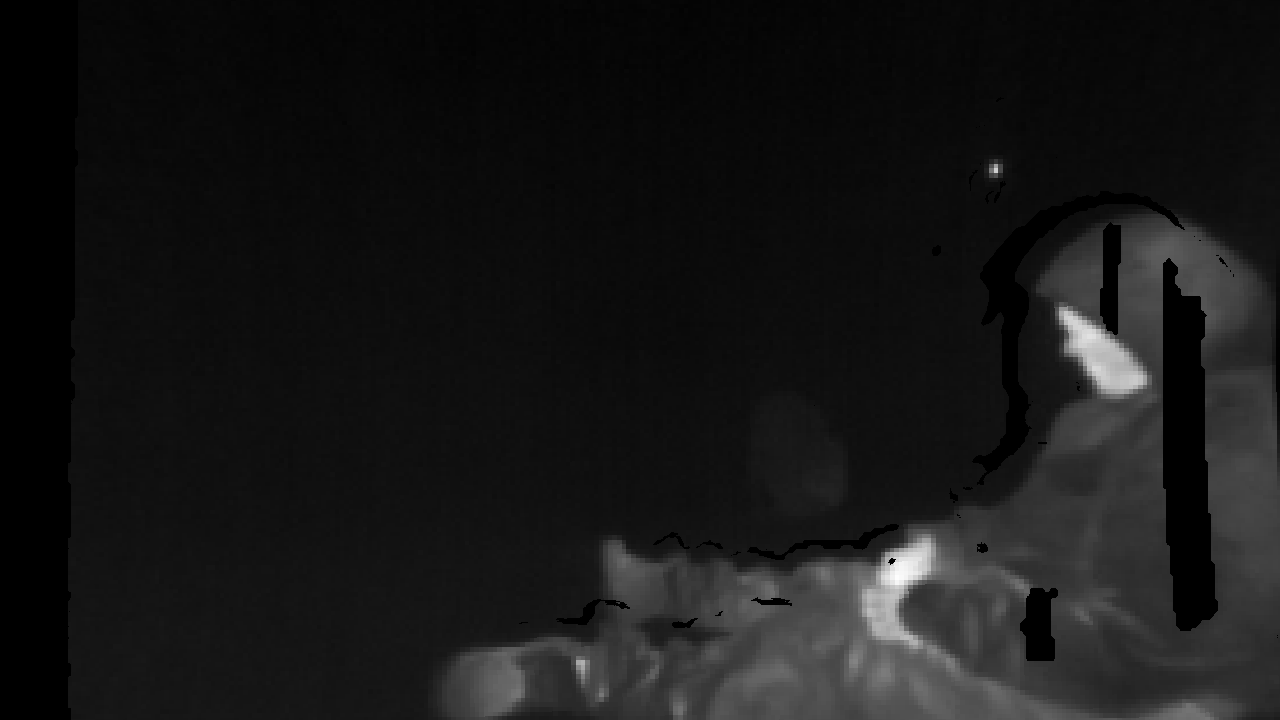} 
        \end{subfigure}
        \begin{subfigure}{2.5cm}
        \includegraphics[width=2.5cm]{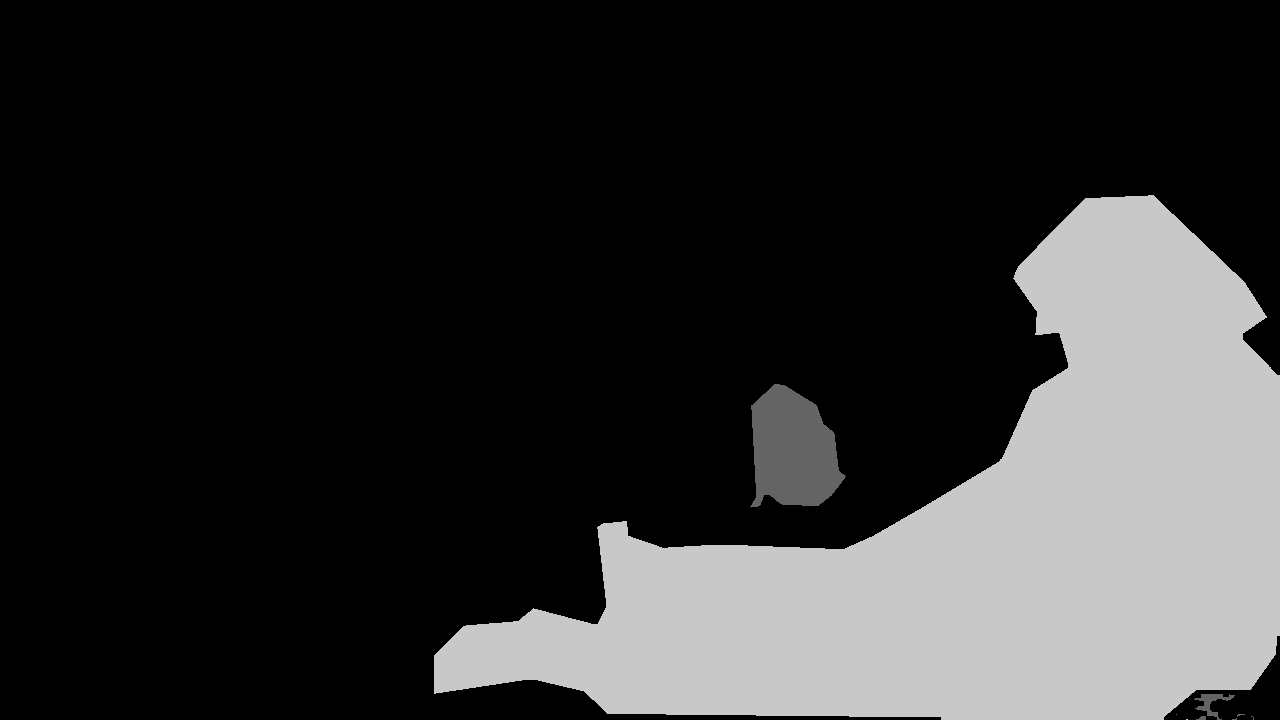} 
        \end{subfigure}
        \begin{subfigure}{2.5cm}
        \includegraphics[width=2.5cm]{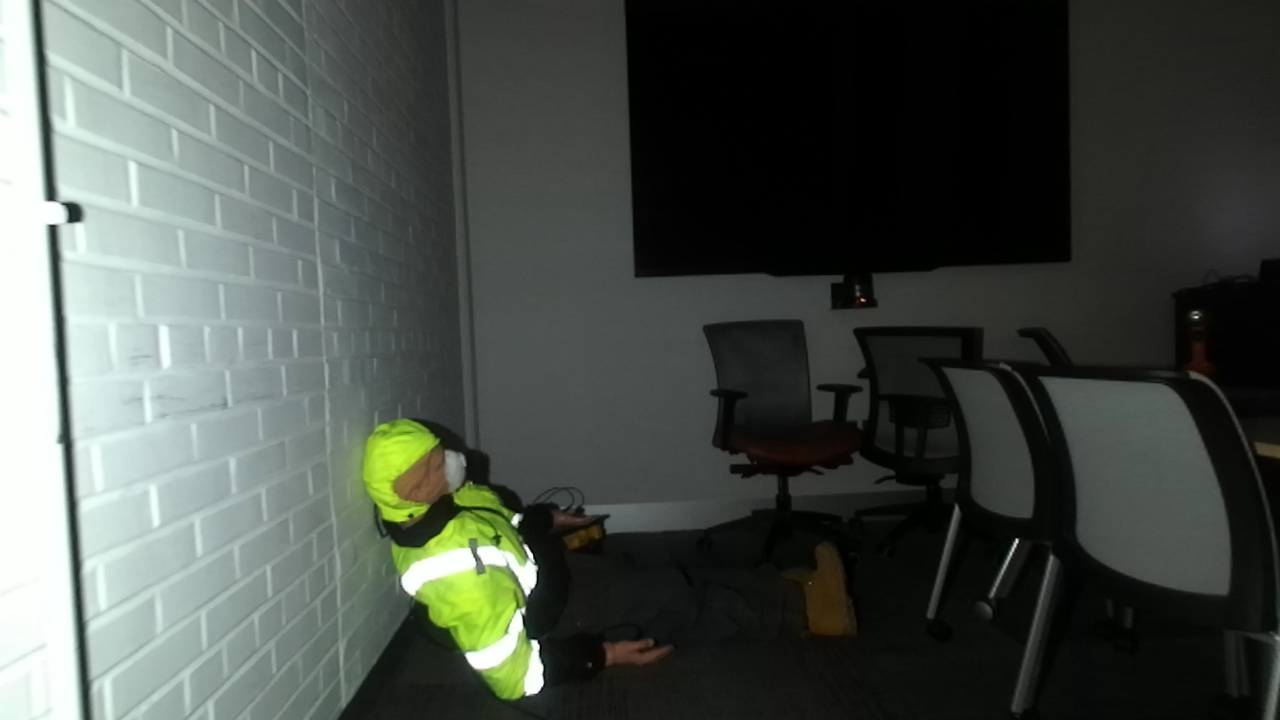} 
        \end{subfigure}
        \begin{subfigure}{2.5cm}
        \includegraphics[width=2.5cm]{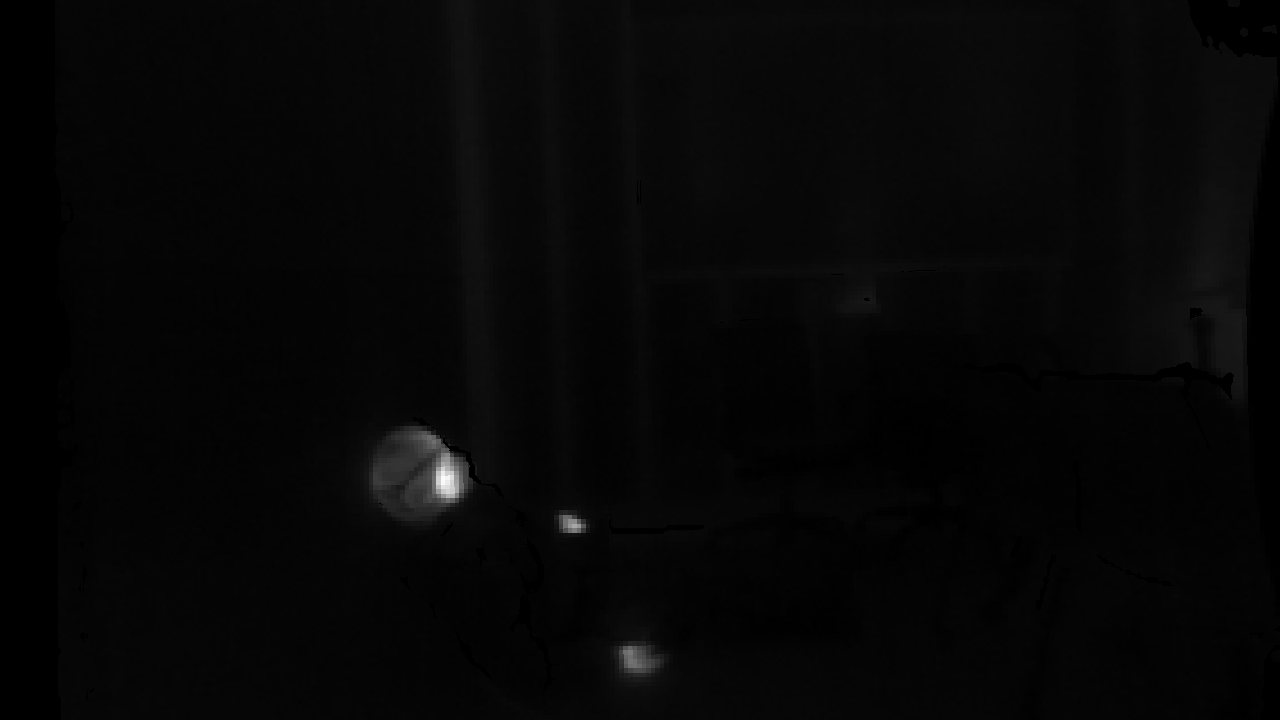} 
        \end{subfigure}
        \begin{subfigure}{2.5cm}
        \includegraphics[width=2.5cm]{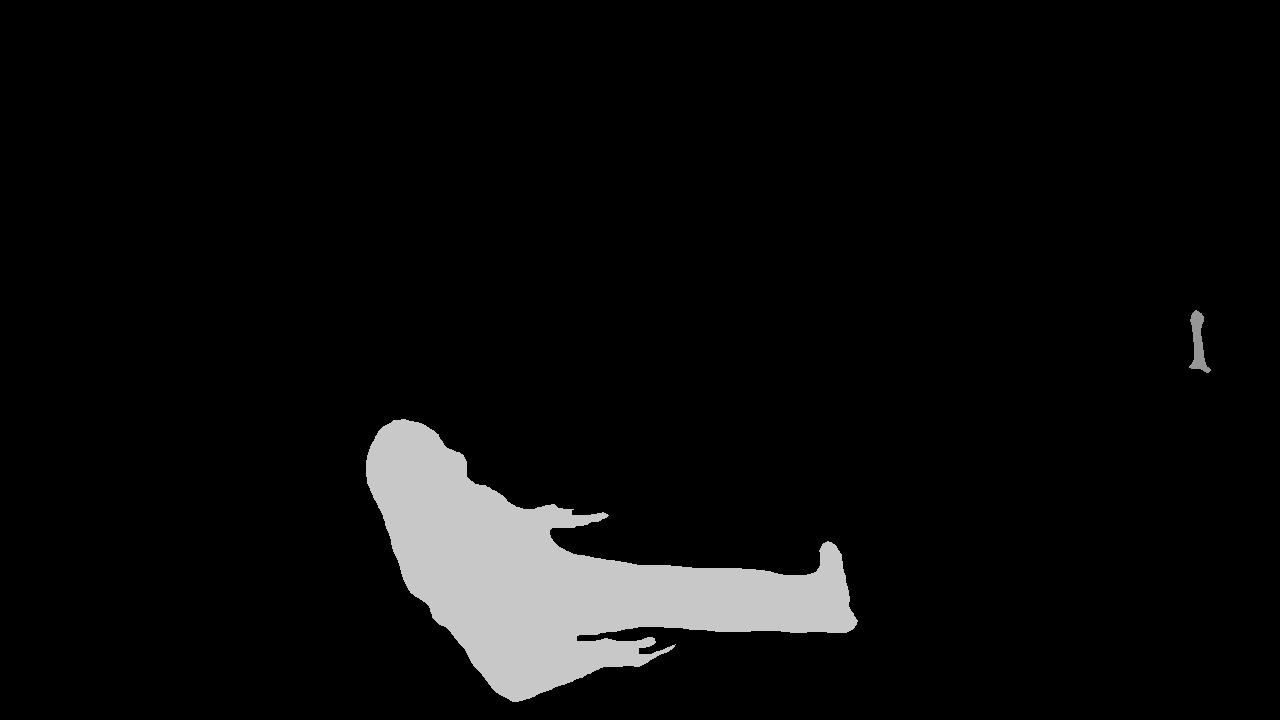} 
        \end{subfigure}
        \begin{subfigure}{2.5cm}
        \includegraphics[width=2.5cm]{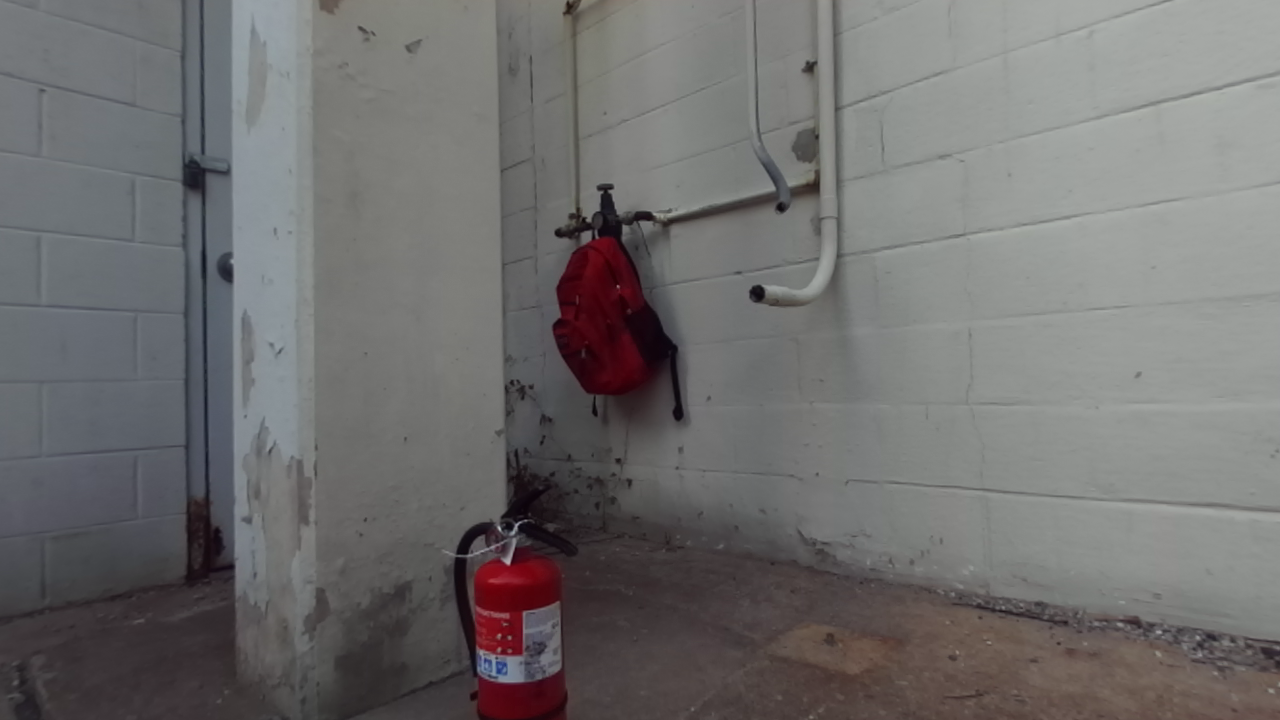} 
        \end{subfigure}
        \begin{subfigure}{2.5cm}
        \includegraphics[width=2.5cm]{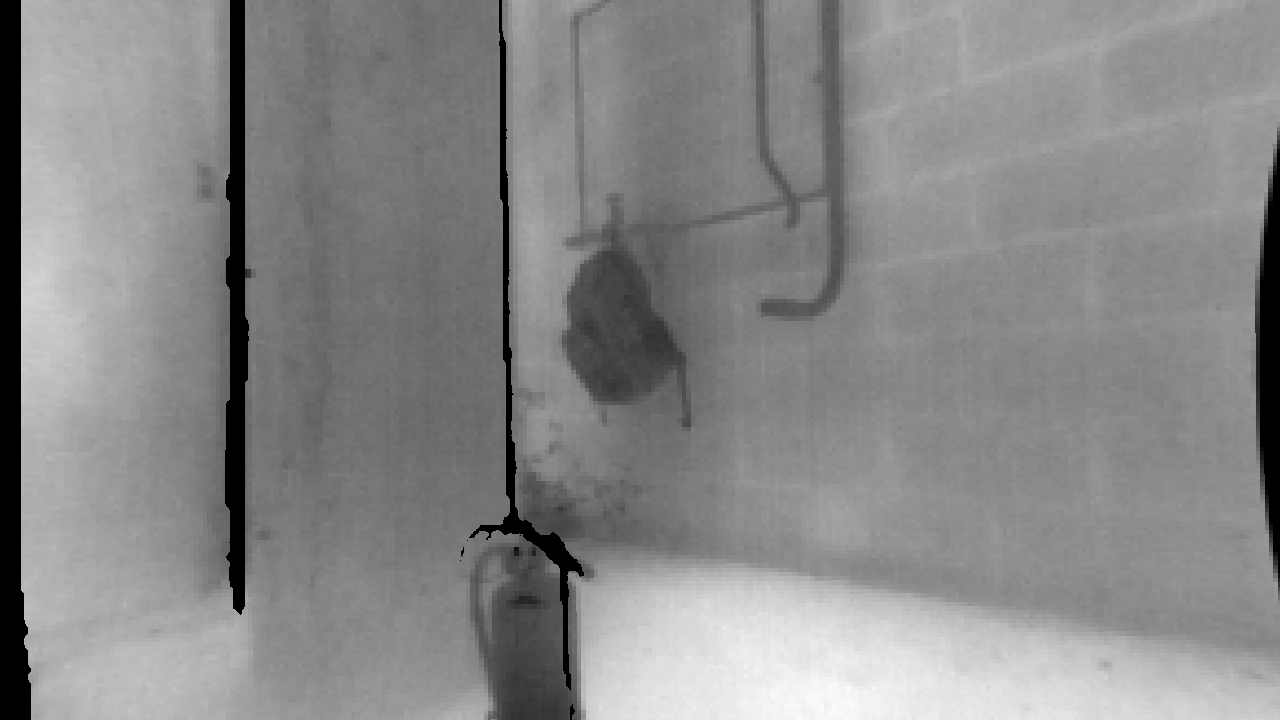} 
        \end{subfigure}
        \begin{subfigure}{2.5cm}
        \includegraphics[width=2.5cm]{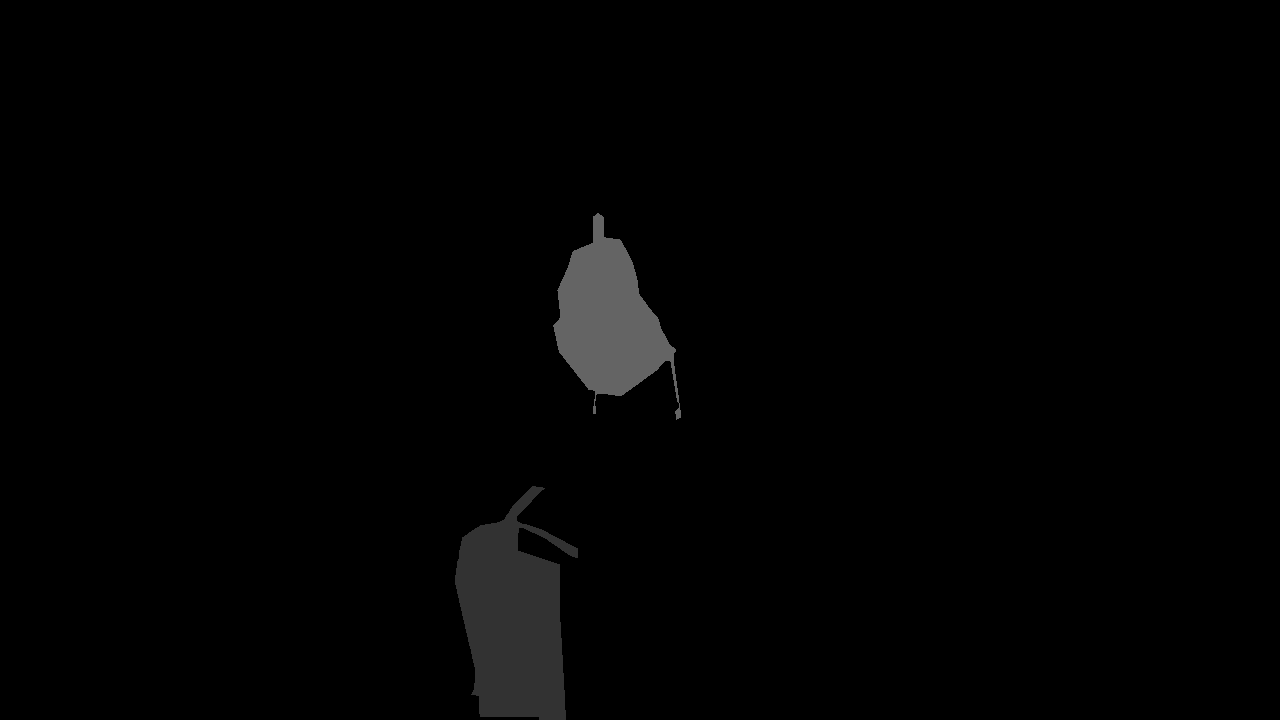} 
        \end{subfigure}
        \caption{RGB, thermal and annotation images from the PST900 dataset \cite{shivakumar2020pst900}}
        \label{fig:pst900}
\end{figure}

Shivakumar et al. introduced Penn Subterranean Thermal 900 Dataset (PST900) \cite{shivakumar2020pst900} containing 894 aligned RGB-T image pairs with ground truth annotations. A Stereolabs ZED Mini stereo camera and a FLIRBoson 320 camera are used for data collection. The PST900 aims to meet the needs of the DARPA Subterranean Challenge\footnote{https://www.subtchallenge.com/} that requires the identification of four objects (fire extinguisher, backpack, hand drill, and thermal mannequin or person) and robustness in various underground situations. Therefore, images are gathered from diverse environments with varying degrees of lighting. Two RGB-T image pairs and the ground truth annotations from the dataset can be seen in Figure \ref{fig:pst900}. Additionally, the dataset provides 3416 annotated RGB images. 


The Freiburg Thermal Dataset \cite{vertens2020heatnet} includes 12051 daytime and 8596 nighttime time-synchronized RGB-T image pairs captured in rural and urban environments. A stereo RGB camera rig (FLIR Blackfly 23S3C) and a stereo thermal camera rig (FLIR ADK) are used for data collection. However, only the testing set including 32 daytime and 32 nighttime images annotated with the following classes: road, sidewalk, building, curb, fence, pole/signs, vegetation, terrain, sky, person/rider, car/truck/bus/train, bicycle/motorcycle and background. 

Multi-model multi-stage network (MMNet) \cite{lan2021mmnet} compares its nighttime segmentation performance with other models using its own dataset. The dataset contains 541 urban scenes RGB and thermal images taken only at night. All the images have a resolution of 300x400. Since the dataset is not publicly available, its use is limited to the MMNet. 

\subsection{Multi-spectral Semantic Segmentation Methods}
\begin{figure*}[h]
    \centering
    \includegraphics[width=\textwidth]{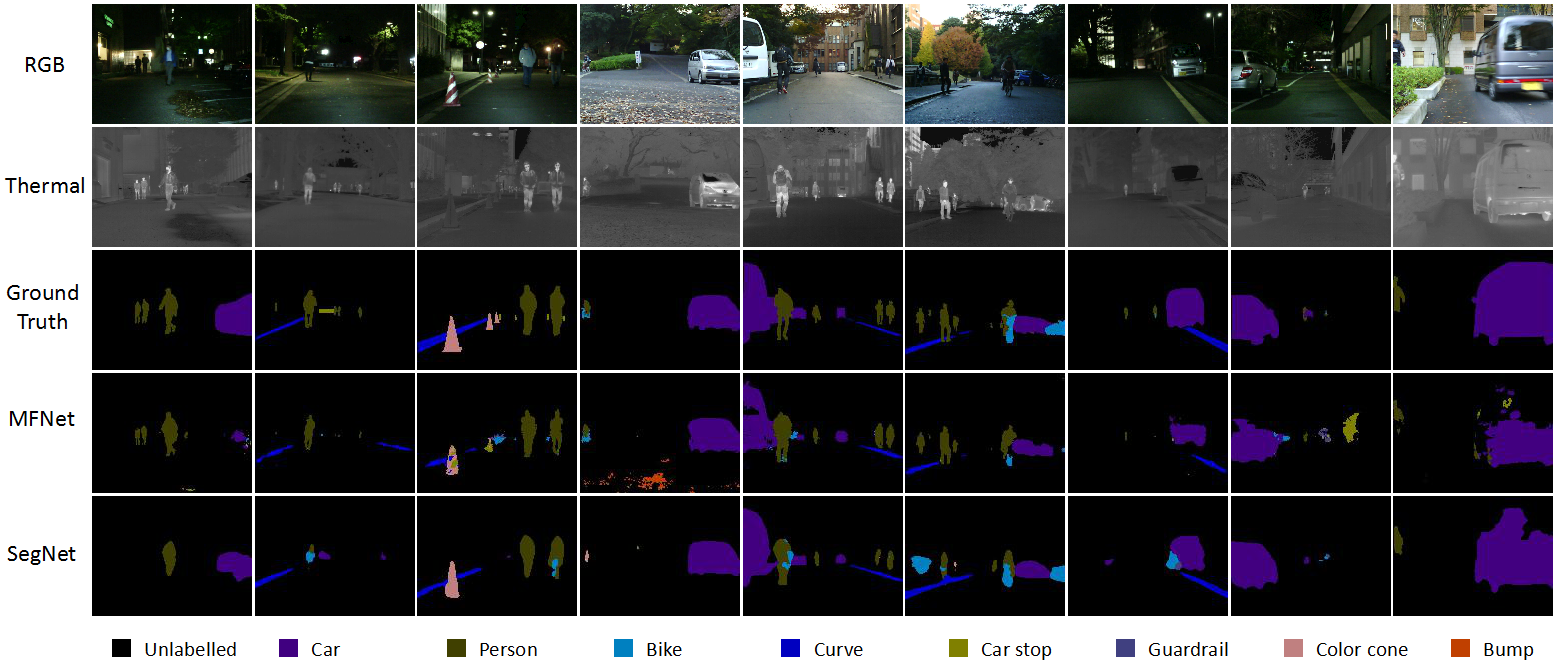} 
    \caption{Some prediction results of MFNet \cite{ha2017mfnet} and Segnet \cite{badrinarayanan2017segnet} on the MFNet dataset \cite{ha2017mfnet}}
    \label{fig:mfnet}
\end{figure*}

Ha et al.\cite{ha2017mfnet} proposed Multi-Spectral Fusion Networks (MFNet) having two identical encoders for thermal and RGB images and one decoder block. Also, the encoder has a mini-inception block with dilated convolution so that the size of the receptive field is enlarged while the time complexity is the same with a normal 3x3 convolutional layer when the number of input and output channels are the same. MFNet aims to achieve high inference speed for real-time semantic segmentation for autonomous vehicles, and MFNet dataset, including RGB-Thermal (RGB-T) urban scene images, is introduced with pixel-level annotations for the self-driving task. MFNet includes a small decoder designed to reduce the number of parameters, and the decoder makes use of the low-level feature maps extracted in encoders to improve up-sampling efficiency. A concatenation operation fuses the outputs of the RGB and infrared (IR) encoders, and the decoder receives the fused result as input. Some segmentation predictions of MFNet and SegNet \cite{badrinarayanan2017segnet} can be seen in Figure \ref{fig:mfnet} which is directly taken from \cite{ha2017mfnet}. 

Sun et al. proposed RGB-Thermal Fusion Network (RTFNet) \cite{sun2019rtfnet}  to achieve semantic segmentation of urban scenes for autonomous vehicles. RTFNet adopts an encoder-decoder structure with two encoders for extracting features of RGB and IR inputs and one decoder restoring the resolution of feature maps. The encoders are identical except the first layers' input channel numbers and slightly changed ResNet-50 \cite{he2016deep} is employed as feature extractors. The infrared feature maps are fused into the RGB encoder through the element-wise summing. The decoder uses the output of the last fusion layer as input to obtain dense predictions. The encoder and decoder of the model are designed asymmetrically, two large encoders and a small decoder. Each decoder layer has two sub-blocks introduced by RTFNet, namely Upception A and Upception B. Upception A does not change the resolution and channel number, whereas Upception B changes, and the final channel number equals the number of classes. Also, Upception blocks have short-cut connections. In short, the decoder block gradually restores the resolution.

The Penn Subterranean Thermal Network (PSTNet) \cite{shivakumar2020pst900} includes independent RGB and Fusion streams to generate segmentation maps from RGB and thermal images. RGB stream can be trained without thermal data since collecting aligned RGB-T images is challenging. Therefore, the designed model uses thermal images to improve the initial segmentation result in the Fusion stream. The RGB stream is a ResNet-18 \cite{he2016deep} architecture with an encoder-decoder and skip-connection scheme similar to U-Net \cite{ronneberger2015u}. The RGB stream is trained with the annotated RGB images to get the per-pixel confidence volume for the classes. This volume, thermal, and RGB input images are concatenated, and the result is passed to the Fusion stream, which is essentially an ERFNet-based \cite{romera2017erfnet} encoder-decoder architecture.

Sun et al. proposed FuseSeg \cite{sun2020fuseseg} employing encoder-decoder structure and two-stage fusion strategy to achieve segmentation in urban scenes. There are two encoders taking three-channel RGB and one-channel thermal images as inputs, and DenseNet-161 \cite{huang2017densely} is employed as the backbone of the encoders. Moreover, FuseSeg introduces a decoder including three modules: a feature extractor with two convolutional layers, an upsampler, and an out block. The upsampler and the out block each have a transposed convolutional layer. The feature extractor is responsible for extracting features from the fused feature maps while keeping the resolution of the feature maps unchanged. The upsampler and the out block increase the resolution by 2. The out block outputs the final segmentation result. Sun et al. also proposed a two-stage fusion strategy to effectively use the multi-spectral inputs and reduce the loss of spatial information due to downsampling. In the first stage of the fusion, feature maps extracted from the inputs in the encoder are fused with element-wise summation in the RGB encoder. The summations are again fused with the corresponding feature maps in the decoder through concatenation.

Vertens et al. \cite{vertens2020heatnet} proposed HeatNet intending to achieve daytime and nighttime image segmentation tasks without costly annotations of nighttime images. The PSPNet \cite{zhao2017pyramid} is exploited as a teacher model to get the annotations of daytime images in the Freiburg Thermal dataset \cite{vertens2020heatnet}. For this purpose, PSPNet is trained on the Mapillary Vistas dataset \cite{neuhold2017mapillary}. Then, a multimodal semantic segmentation network is trained using RGB and thermal daytime images annotated by the teacher model. The multimodel network also exploits PSPNet architecture and the first two block of the corresponding ResNet-50 \cite{he2016deep} encoder. Moreover,  a domain adaptation method similar to \cite{tsai2018learning} is proposed to obtain nighttime segmentation results; therefore, a domain discriminator is employed after the softmax layer of the multimodal RGB-T model. Besides, the training is conducted using an alternating training scheme.

Graded-Feature Multilabel-Learning Network (GMNet) \cite{zhou2021gmnet} includes two encoders for feature extraction and three grading decoding stages to restore original resolution. The proposed model employed ResNet-50 \cite{he2016deep} as the backbone of the encoders. The fully connected and average pooling layers of ResNet are removed as they may result in the loss of spatial information and details. GMNet divides multilevel features into senior, intermediate, and junior grades. The features extracted from the ResNet's last three layers, in which the visual receptive fields are enlarged, are selected as senior features. Besides, the features from the first layer, which have more detailed information, are selected as junior features. Moreover, GMNet introduces two fusion modules, the shallow feature fusion module (SFFM) and the deep feature fusion module (DFFM), to use the junior, intermediate, and senior features. SFFM fuses the features from the first two layers of the encoders, whereas DFFM accomplishes the fusion operation for the last three layers. Finally, semantic, binary, and boundary loss functions are used to find the optimum parameters of the model.

Multi-Modal Multi-Stage Network (MMNet) \cite{lan2021mmnet} tackles the semantic segmentation problem by employing three encoder-decoder structures. In the first stage of the network, two separate encoder-decoder structures process RGB and thermal images with no information interactions between the modalities. In the second stage, one encoder-decoder fuses and refines the features from the first stage. The proposed model deploys ResNet-18 \cite{he2016deep} as encoders in the first stage, whereas Mini Refinement Block (MRB) is proposed as the encoder for the second stage. Each encoder sends information to the corresponding decoder using skip connections. As the direct connection may impact the fusion performance, EFEM (Efficient feature enhancement module) has been proposed to reduce the semantic gap between encoders and decoders.

Zhang et al. \cite{zhang2021abmdrnet} proposed Adaptive-weighted Bi-directional Modality Difference Reduction Network (ABMDRNet) containing three parts: Modality Difference Reduction and Fusion (MDRF) subnetwork, Multi-Scale Spatial Context (MSC) module, and Multi-Scale Channel Context(MCC) module. All RGB-T networks strive to use complementary information from RGB and thermal images to their advantage. The integration and utilization of multi-modality complementary information from RGB and thermal images may be hampered by the modality difference generated by distinct imaging mechanisms. Therefore, the MDRF subnetwork uses a bridging-then-fusing strategy to reduce the modality difference and utilize the multi-modality complementary information. An MSC module and an MCC module are designed to exploit multi-scale contextual knowledge of cross-modality features and their long-range relationships along spatial and channel dimensions.

Xu et al. \cite{xu2021attention} proposed Attention Fusion Network (AFNet) containing an attention fusion module to guide the fusion operation of multi-spectral inputs. Also, AFNet employs two identical encoders for feature extraction and a single decoder for resolution restoration. The encoders are designed based on the ResNet-50 \cite{he2016deep} with dilated convolutions. Also, the downsampling operations in the last two residual blocks in ResNet are removed. To make full use of the complementary properties of the RGB and thermal images, AFNet designed an attention fusion module. The attention matrices are obtained considering the cross-spectral and the global contextual relations of the images. The fusion operation takes place under the guidance of attention matrices, and the decoder uses the fused feature map. Moreover, the decoder employs three interpolations and three convolutional layers to obtain the segmentation result.

\subsection{Analysis \& Results}

Exploiting thermal images as well as RGB images may improve the segmentation network in terms of accuracy and robustness. For multi-spectral input semantic segmentation, several methods have been developed, and this review describes the similarities and differences of these methods in many aspects. The proposed architectures and fusion strategies tackle the difficulties of employing two images and producing a precise segmentation result. Two encoders and a single decoder are commonly used in RGB-T methods, such as \cite{ha2017mfnet}, \cite{sun2019rtfnet}, \cite{sun2020fuseseg}, \cite{zhou2021gmnet}, and \cite{xu2021attention} to extract features and restore the resolution. Moreover, \cite{shivakumar2020pst900} employs two distinct streams to process RGB images and the fused features, respectively. In this way, it can be trained by using only RGB images, and thermal images may provide further improvement. Also, \cite{lan2021mmnet} has three encoder-decoder structures with EFEM connections. \cite{vertens2020heatnet} achieves nighttime image segmentation as well as daytime image segmentation by using a domain adaptation method. Furthermore, different fusion strategies are proposed to use complementary information from different modalities without information conflicts. \cite{ha2017mfnet} concatenates the outputs of the encoders, whereas the \cite{sun2019rtfnet} fuses the thermal features into the RGB encoder through element-wise summation. Besides, more complex fusion strategies are proposed for the fusion operation, such as two-stage fusion, bridging-then-fusing strategies, attention fusion module, SFFM, and DFFM. 

The MFNet dataset \cite{ha2017mfnet} includes both day and night RGB images with aligned thermal images. Since the images in the dataset can provide complementary information, RGB-T correlations are essential. The quantitative results of the RGB-T methods on test images from \cite{ha2017mfnet} can be found in Table \ref{table:mfnet_results}. The results are shown using a standard evaluation metric, mean Intersection over Union (mIoU). All the results provided in the table are taken from the original papers of the mentioned methods. On the other hand, the PST900 dataset \cite{shivakumar2020pst900} is more challenging for thermal fusion networks since plenty of information is provided by RGB alone, and the same object images are captured at both above and below the ambient temperature, making learning RGB-T correlations challenging. The PSTNet \cite{shivakumar2020pst900} reports better results on PST900 because the model has a separate RGB stream and employs a late fusion approach.

According to the revealed results in \cite{vertens2020heatnet}, on the MFNet dataset with three classes (person, car, bicycle), \cite{vertens2020heatnet} and \cite{ha2017mfnet} has comparable results, while \cite{sun2019rtfnet} outperforms with 0.707 mIoU. Also, the paper indicates that \cite{vertens2020heatnet} achieves mIoU score of 0.597 on the Freiburg Thermal dataset while \cite{ha2017mfnet} and \cite{sun2019rtfnet} perform with only 0.314 mIoU and 0.586 mIoU, respectively.

The inference speed of \cite{ha2017mfnet}, RTFNet-50, RTFNet-152 \cite{sun2019rtfnet} and \cite{sun2020fuseseg} are declared as 229.86, 88.87, 34.07 and 30.01 FPS according to the results announced in \cite{sun2020fuseseg}. Also, \cite{lan2021mmnet} and \cite{xu2021attention} reports their inference speeds slightly higher than \cite{sun2019rtfnet}.

\begin{table}[!h]
\centering
\caption{The Results of the RGB-T Methods on MFNet Dataset \cite{ha2017mfnet}}
\resizebox{0.35\textwidth}{!}
{%
\renewcommand{\arraystretch}{0.7}

\begin{tabular}{ll}
\hline
\textbf{\tiny{Method}} & \textbf{\tiny{mean IoU}} \\ \hline
\tiny{MFNet \cite{ha2017mfnet}}  & \tiny{0.649} \\ \hline
\tiny{RTFNet-50 \cite{sun2019rtfnet}}  & \tiny{0.517} \\ \hline
\tiny{RTFNet-152 \cite{sun2019rtfnet}}  & \tiny{0.532} \\ \hline
\tiny{PSTNet \cite{shivakumar2020pst900}}  & \tiny{0.484} \\ \hline
\tiny{FuseSeg \cite{sun2020fuseseg}}  & \tiny{0.545} \\ \hline
\tiny{GMNet \cite{zhou2021gmnet}}  & \tiny{0.573} \\ \hline
\tiny{MMNet \cite{lan2021mmnet}}  & \tiny{0.528} \\ \hline
\tiny{ABMDRNet \cite{zhang2021abmdrnet}}  & \tiny{0.548} \\ \hline
\tiny{AFNet \cite{xu2021attention}}  & \tiny{0.546} \\ \hline
\end{tabular}
}
\label{table:mfnet_results}
\end{table}

\section{Semantic Segmentation Using Only Infrared Spectrum}
In terms of capturing details under adverse environmental conditions, thermal imaging cameras outperform visual imaging cameras. Thermal imaging cameras are widely used in the defense industry, and since they have become more affordable, they have gained popularity in various other applications. Therefore, a couple of approaches have been developed to achieve high accuracy under a wide range of conditions by using only infrared images. Unlike methods using RGB-T image pairs, aligned images are not required, making data collection easier. But still, the lack of a thermal image dataset limits the number of works conducted in the area. Also, extracting features might be challenging due to thermal crossover, low resolution, and contrast of the infrared images.

\subsection{Thermal Datasets}
\begin{figure}[h]
    \centering
        \begin{subfigure}{3cm}
        \includegraphics[width=3cm]{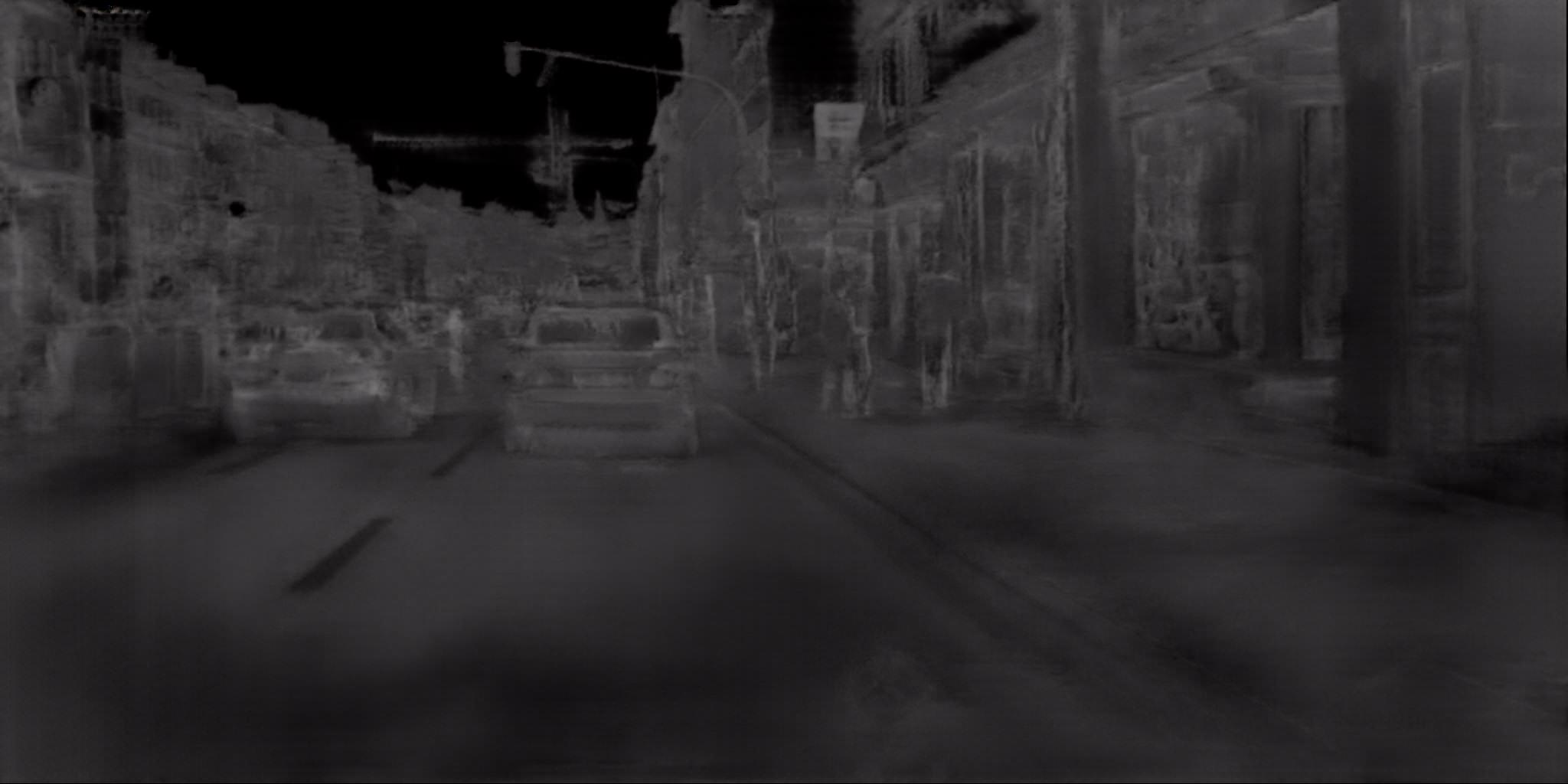} 
        \end{subfigure}
        \begin{subfigure}{3cm}
        \includegraphics[width=3cm]{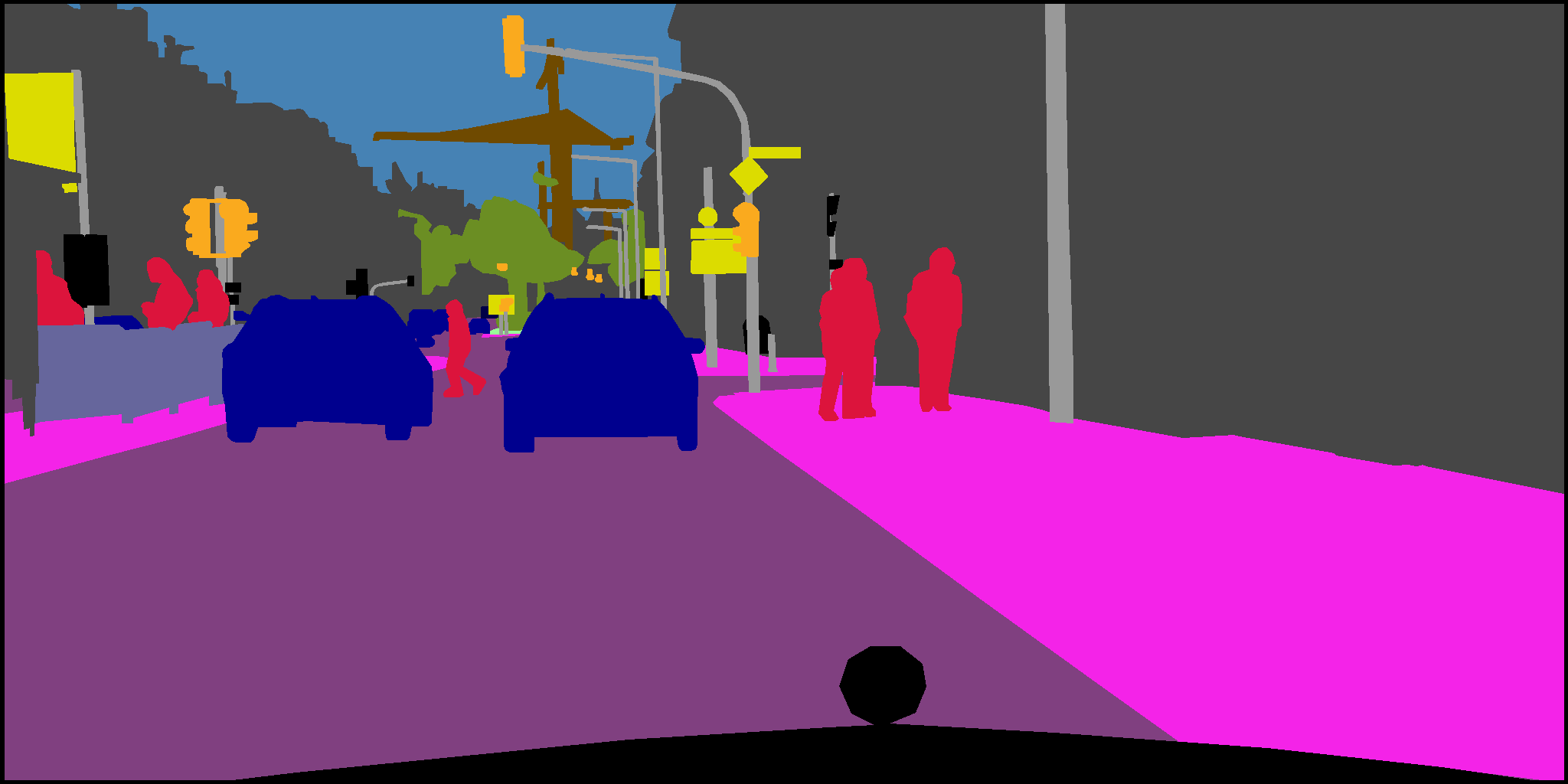}  
        \end{subfigure}
        \begin{subfigure}{3cm}
        \includegraphics[width=3cm]{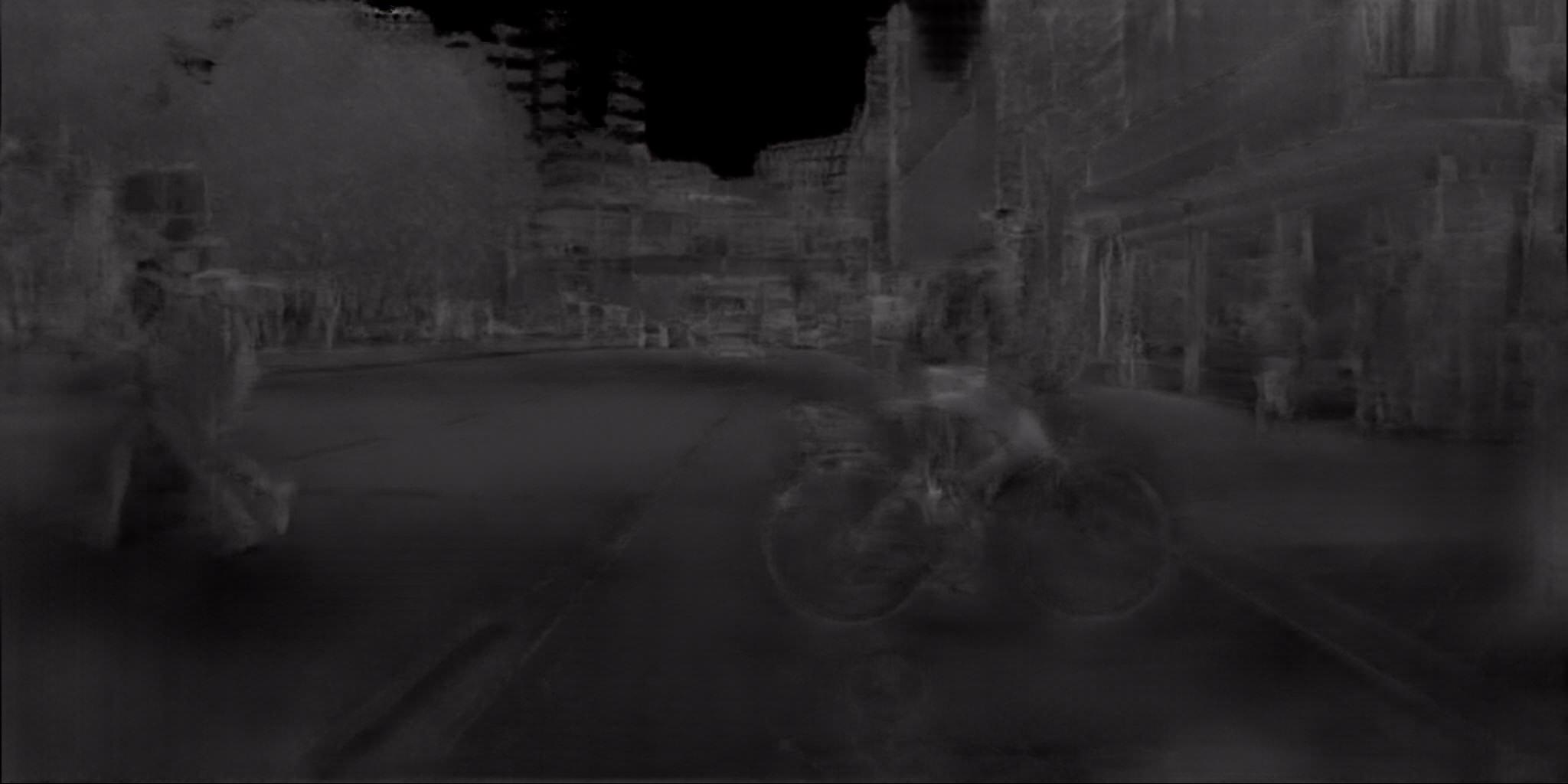} 
        \end{subfigure}
        \begin{subfigure}{3cm}
        \includegraphics[width=3cm]{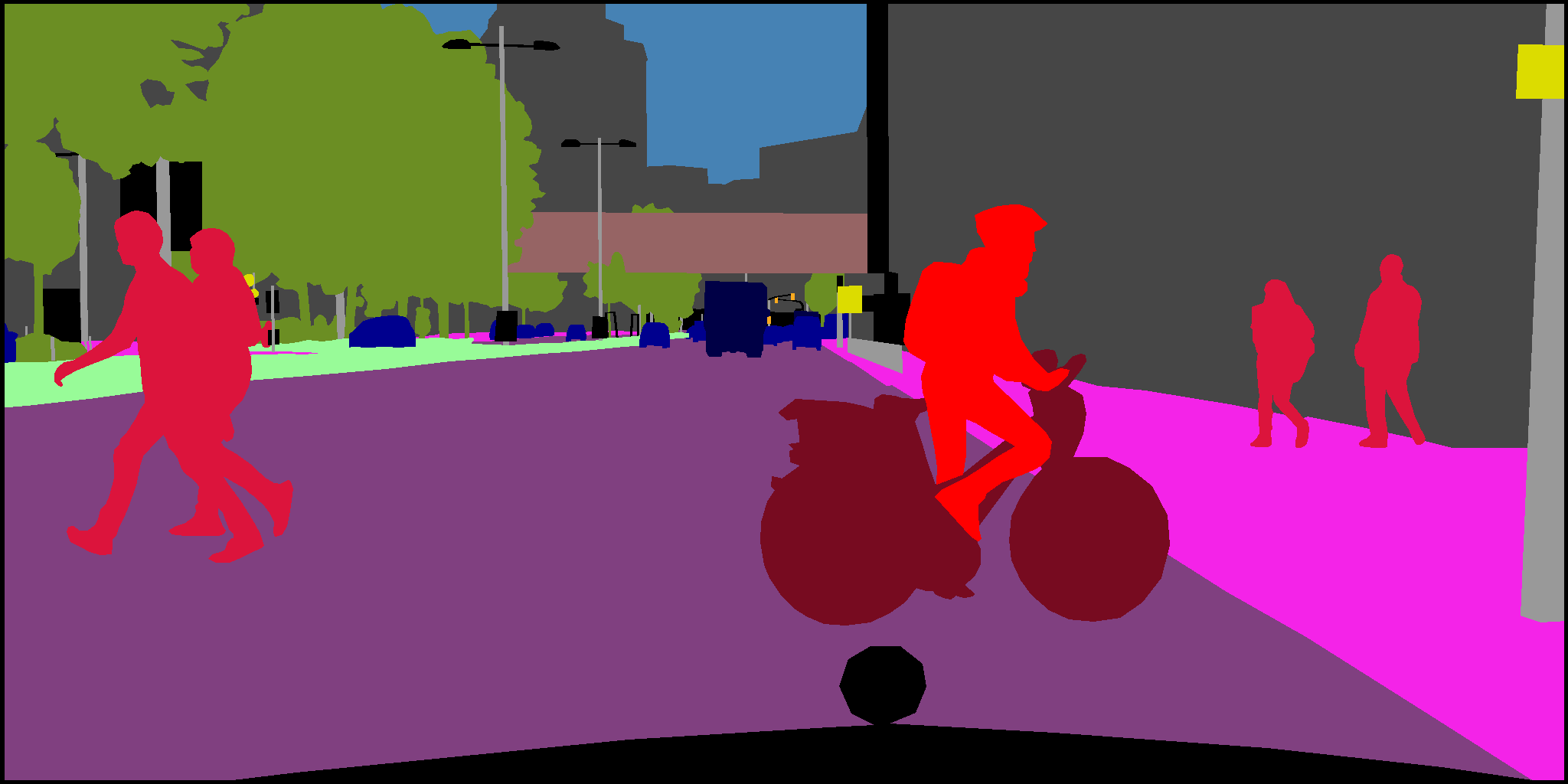}  
        \end{subfigure}
        \begin{subfigure}{3cm}
        \includegraphics[width=3cm]{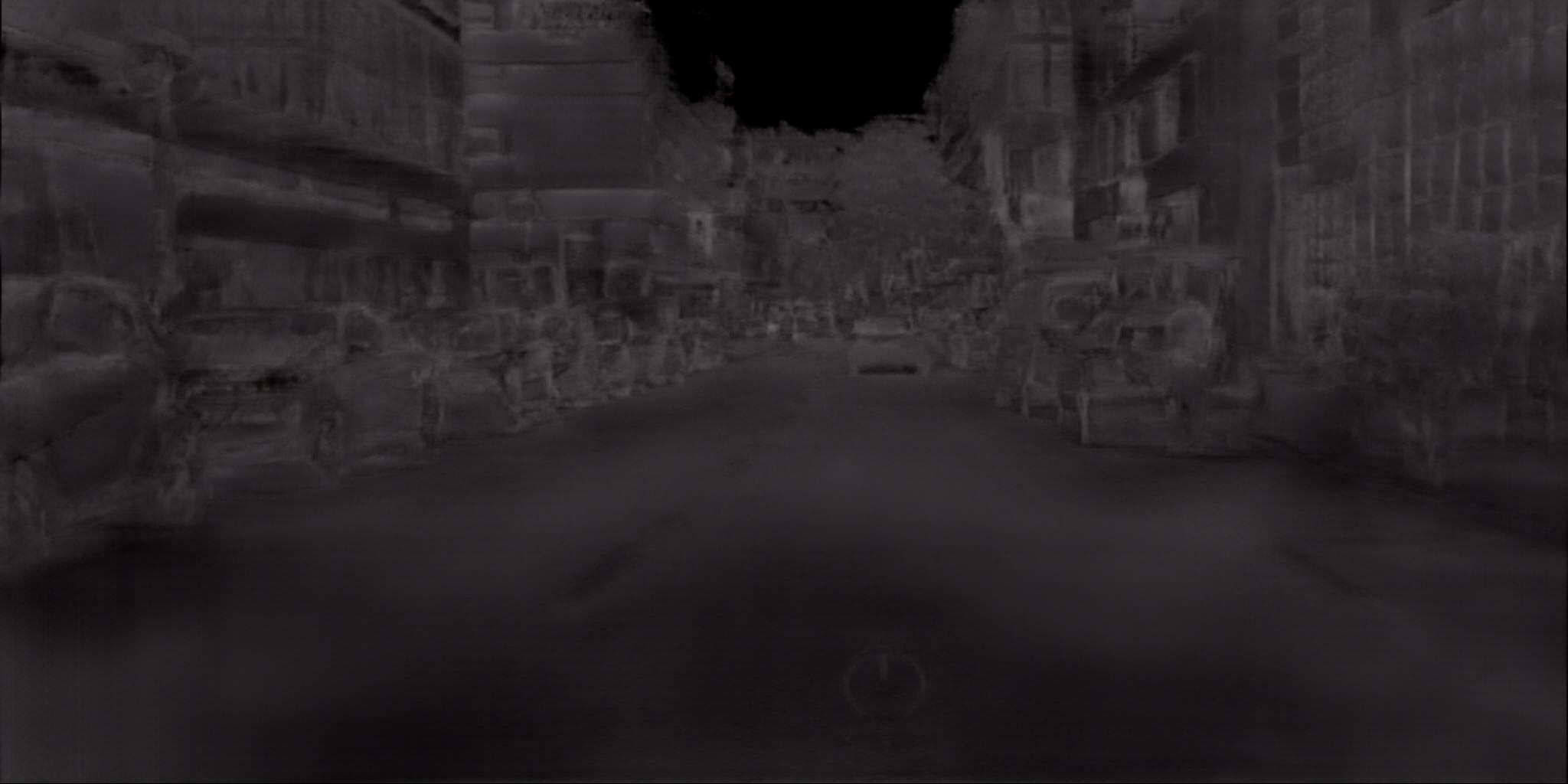} 
        \end{subfigure}
        \begin{subfigure}{3cm}
        \includegraphics[width=3cm]{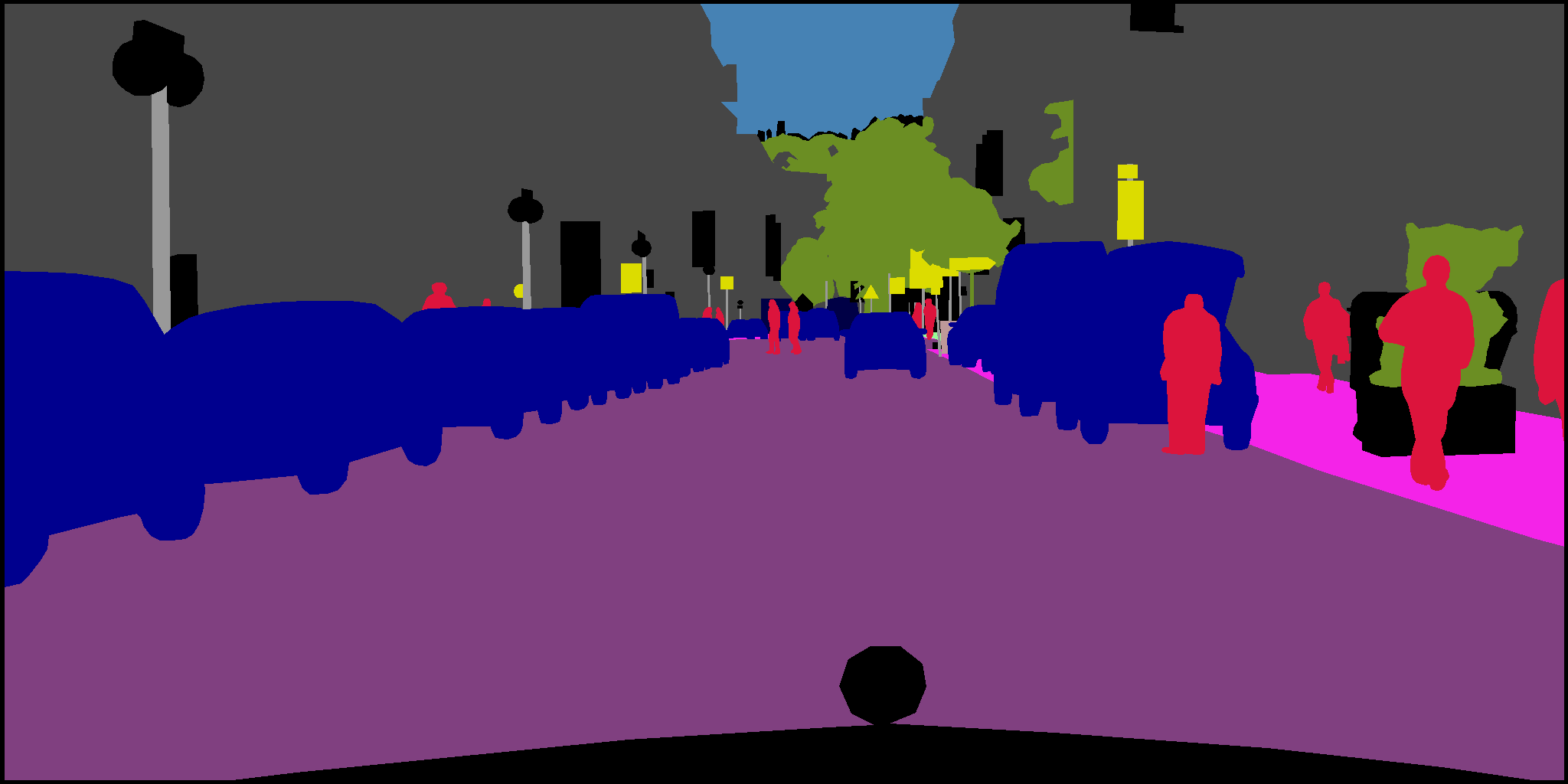}  
        \end{subfigure}
        \caption{Synthetically generated thermal images and ground truth annotations in the SODA dataset \cite{li2020segmenting}}
        \label{fig:soda_syn}
\end{figure}

\begin{figure}[h]
    \centering
        \begin{subfigure}{3cm}
        \includegraphics[width=3cm]{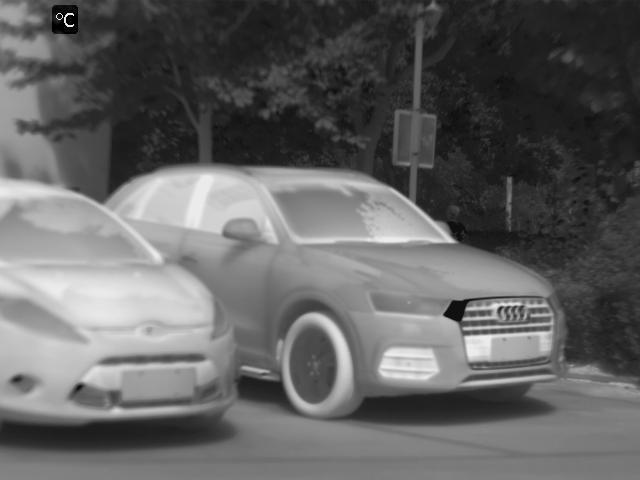} 
        \end{subfigure}
        \begin{subfigure}{3cm}
        \includegraphics[width=3cm]{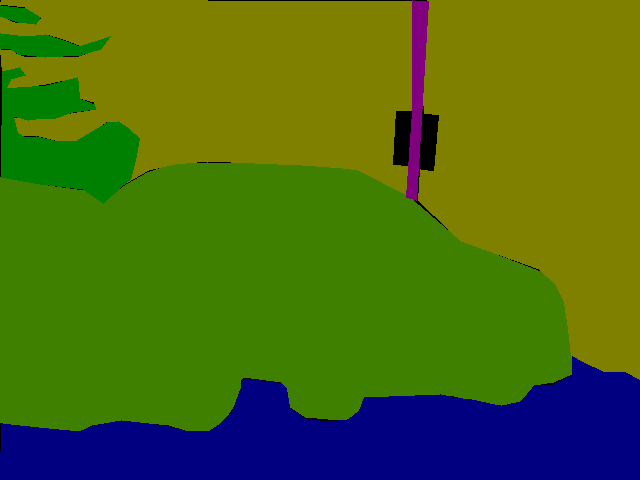}  
        \end{subfigure}
        \begin{subfigure}{3cm}
        \includegraphics[width=3cm]{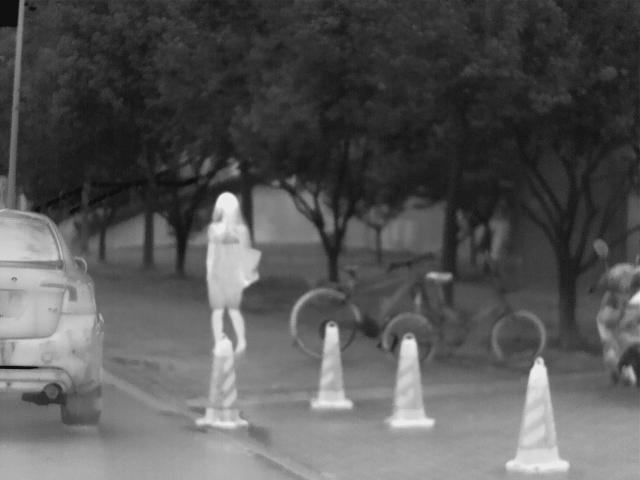} 
        \end{subfigure}
        \begin{subfigure}{3cm}
        \includegraphics[width=3cm]{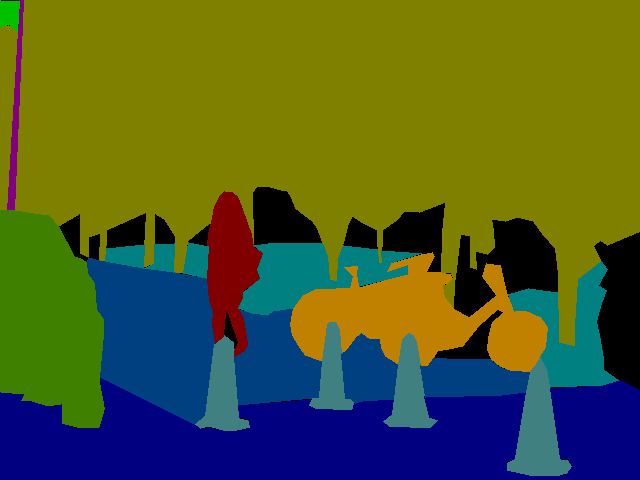}  
        \end{subfigure}
        \begin{subfigure}{3cm}
        \includegraphics[width=3cm]{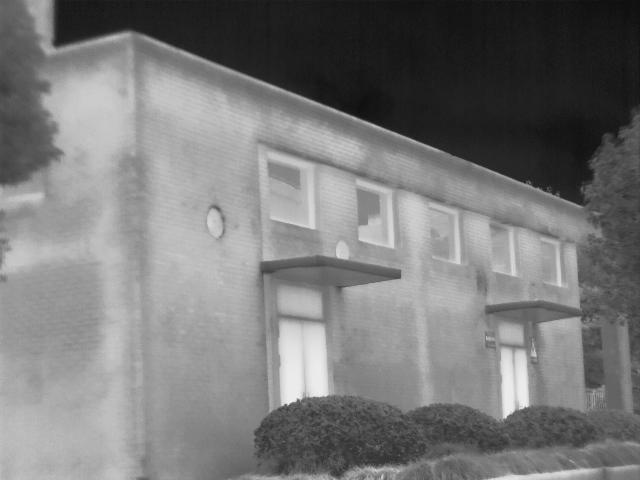} 
        \end{subfigure}
        \begin{subfigure}{3cm}
        \includegraphics[width=3cm]{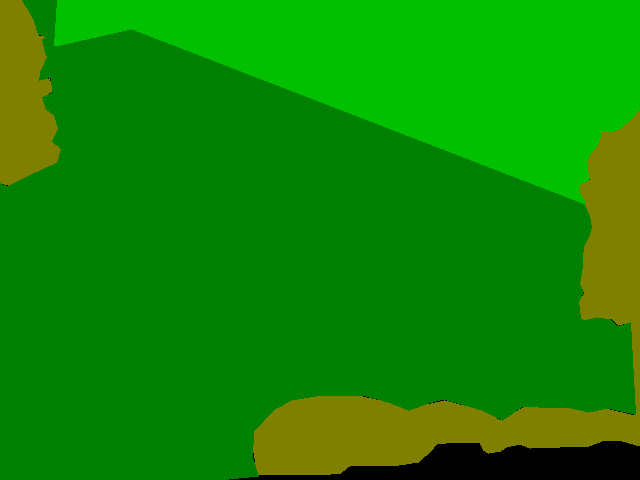}  
        \end{subfigure}
        \caption{Thermal images and ground truth annotations from the real subset of the SODA dataset \cite{li2020segmenting}}
        \label{fig:soda_real}
\end{figure}

Li et al. introduced Segment Objects in Day and Night (SODA) dataset \cite{li2020segmenting}. The SODA consists of 2168 real and 5000 synthetically generated thermal images. The real subset is captured by a FLIR camera (SC260). The thermal images generated from annotated RGB images are included in the synthetic subset. An image-to-image translation method, pix2pixHD \cite{wang2018high}, is trained with KAIST Multispectral Pedestrian Dataset \cite{hwang2015multispectral}. After training the model, the synthetic subset is generated from Cityscapes \cite{cordts2016cityscapes}. Figure \ref{fig:soda_syn} shows some synthetically generated thermal images and ground truth annotations. Labels of the generated thermal images can be obtained directly from RGB annotations. Besides, the real subset images are manually annotated. Three thermal images and annotations from the real subset can be seen in Figure \ref{fig:soda_real}.  

For pedestrian detection from thermal images, there are a few well-known datasets such as OSU Thermal Pedestrian Database (OSUT) \cite{davis2005two}, Terravic Motion IR Database (TMID) \footnote{http://vcipl-okstate.org/pbvs/bench/}, and Pedestrian Infrared/Visible Stereo Video Dataset (PISVD) \cite{bilodeau2014thermal}. However, these datasets are not suited for the segmentation task due to the lack of annotations. Wang et al. \cite{wang2019thermal} introduced a new dataset including thermal pedestrian images from the driver's perspective for autonomous driving applications. The dataset consists of 1031 thermal images at a resolution of 720x480 sampled from 25 scene videos. The dataset is also split into two equal parts for train and test sets. However, the dataset is not publicly available. 

Another application of the thermal semantic segmentation might be the ground vehicle segmentation from aerial images. In this context, NPU\_CS\_UAV\_IR\_DATA \cite{liu2018real} dataset includes UAV-based infrared vehicle images. The dataset also provides four groups of road images for testing. Flying altitude, resolution of the images, and ambient temperature differ in these groups. Also, the captured images differ in terms of the number of vehicles and surroundings. 

For the networks aiming for good segmentation results despite illumination and noise, the Low Illumination Image dataset (LII) \cite{chen2020nv} includes manually labeled thermal, motion blur, night, and weak lighting images. The images' average SNR (signal-to-noise ratio) is 25.5 dB. 

\begin{figure}[h]
    \centering
        \begin{subfigure}{3cm}
        \includegraphics[width=3cm]{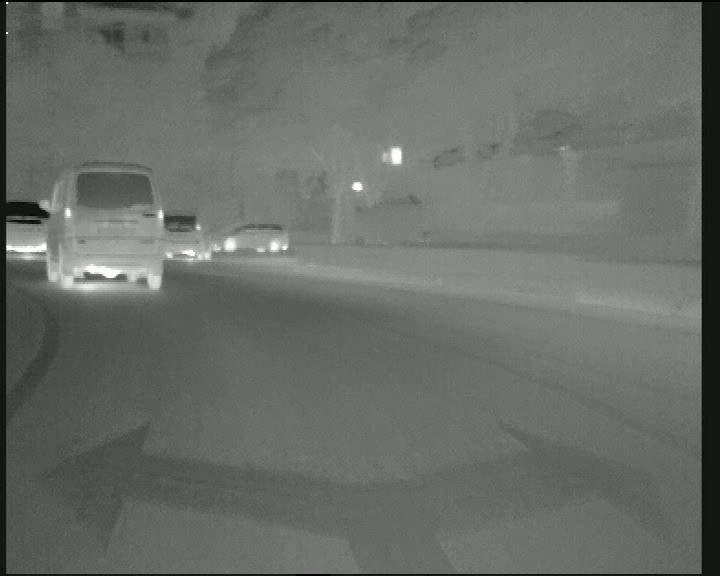} 
        \end{subfigure}
        \begin{subfigure}{3cm}
        \includegraphics[width=3cm]{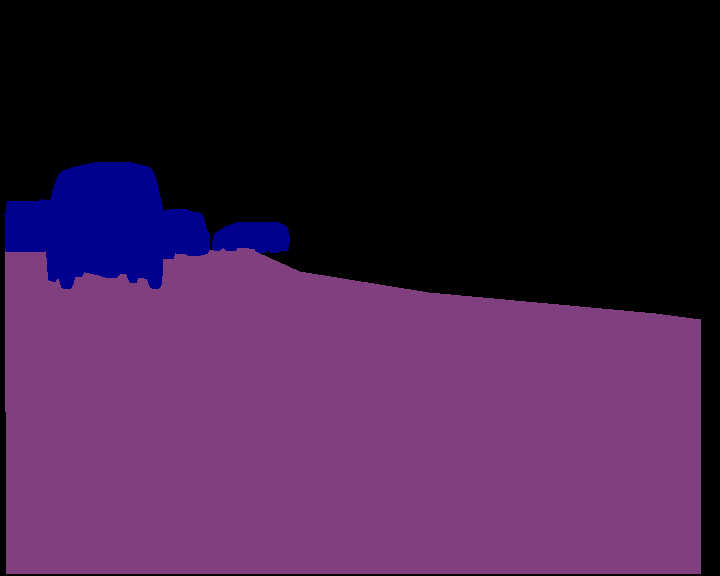}  
        \end{subfigure}
        \begin{subfigure}{3cm}
        \includegraphics[width=3cm]{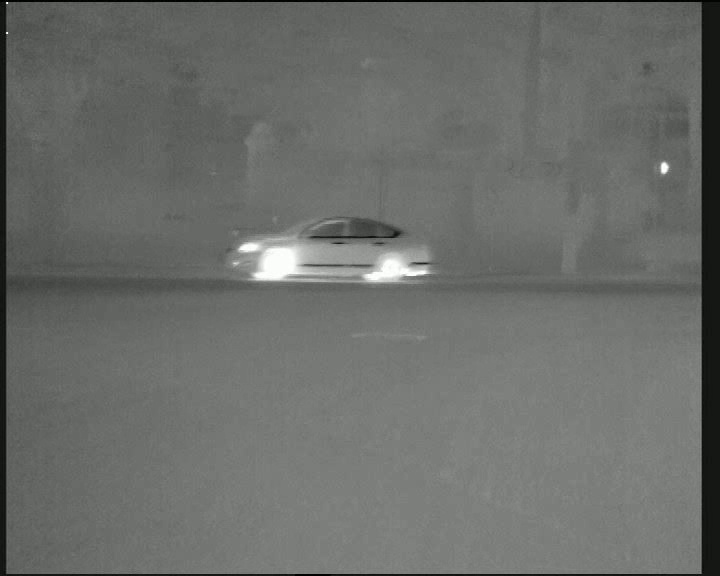} 
        \end{subfigure}
        \begin{subfigure}{3cm}
        \includegraphics[width=3cm]{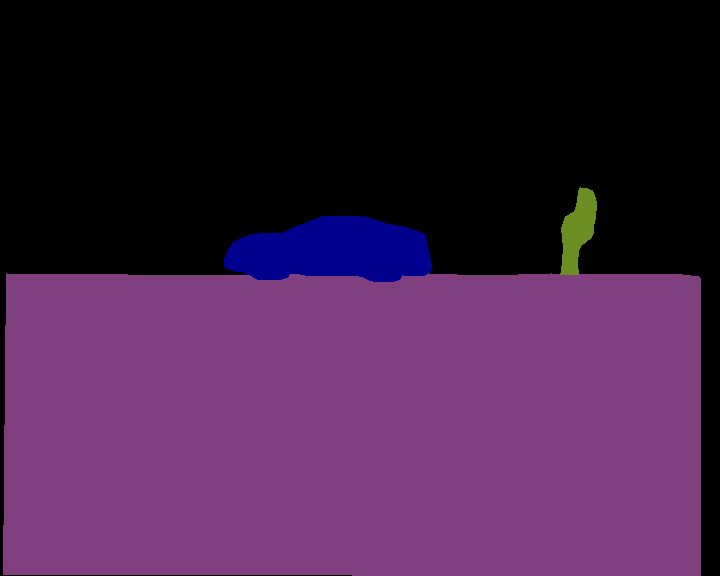}  
        \end{subfigure}
        \begin{subfigure}{3cm}
        \includegraphics[width=3cm]{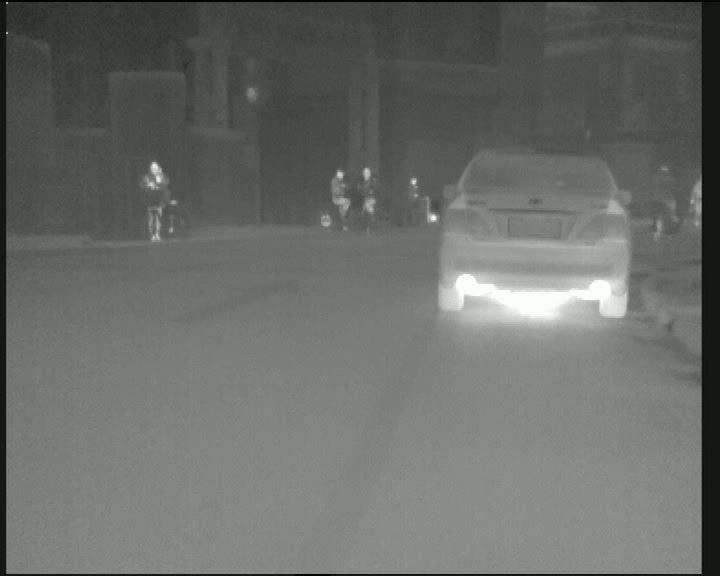} 
        \end{subfigure}
        \begin{subfigure}{3cm}
        \includegraphics[width=3cm]{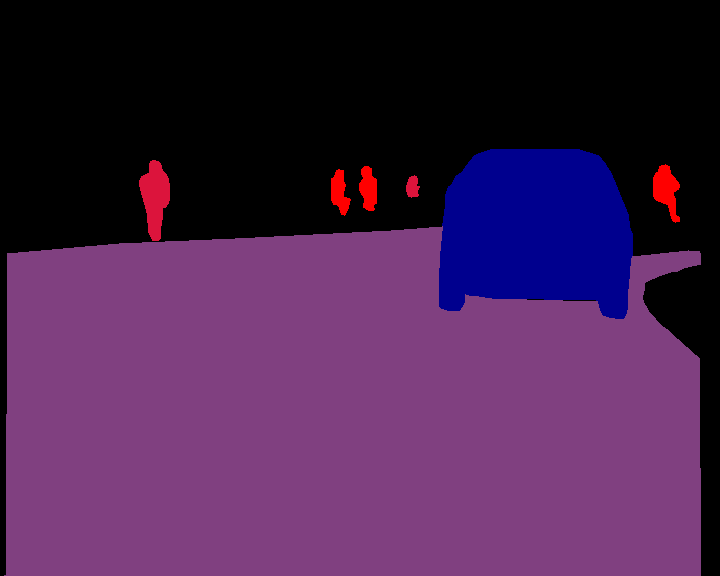}  
        \end{subfigure}
        \begin{subfigure}{3cm}
        \includegraphics[width=3cm]{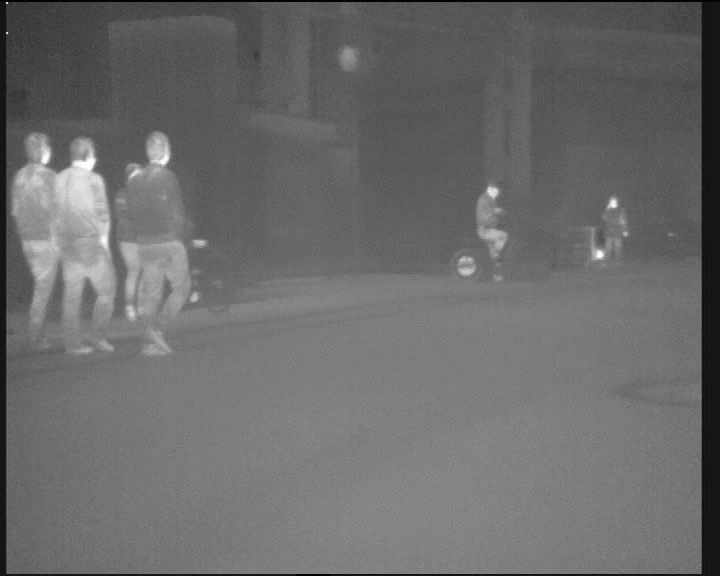} 
        \end{subfigure}
        \begin{subfigure}{3cm}
        \includegraphics[width=3cm]{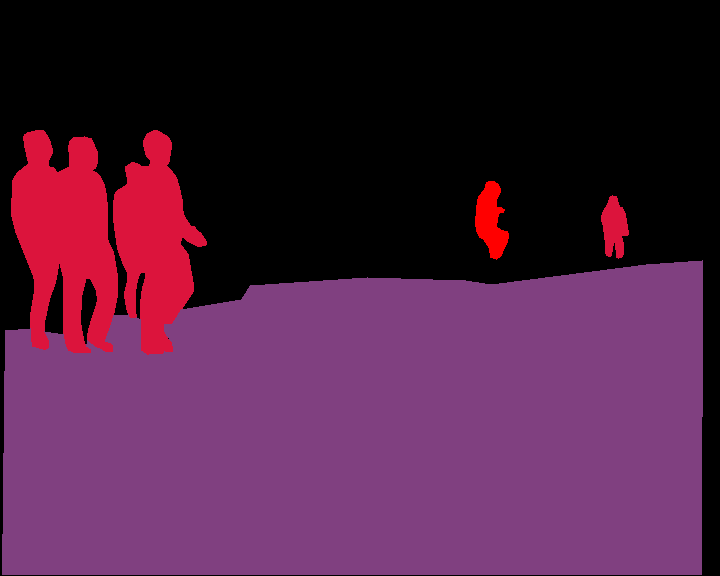}  
        \end{subfigure}
        \caption{Thermal images and corresponding ground truth images from the SCUT-Seg dataset \cite{xiong2021mcnet}}
        \label{fig:scut_seg}
\end{figure}

Xiong et al. introduced SCUT-Seg dataset \cite{xiong2021mcnet} which includes nighttime driving scenes from different environments. The dataset includes 2010 thermal images with semantic-wise annotations for ten classes (background, road, person, rider, car, truck, fence, tree, bus, and pole). Also, instance-wise annotations are provided for future works. The training and testing sets consist of 1365 and 665 images, respectively. Four example images and their ground truth annotations from the training set are presented in Figure \ref{fig:scut_seg}. 

\subsection{Thermal Semantic Segmentation Methods}
Edge-Conditioned CNN (EC-CNN) \cite{li2020segmenting} exploits edge prior information to increase the quality of segmentation output since thermal crossover and thermal sensors cause ambiguous object boundaries and imaging noise, respectively. Some gated feature-wise transform (GFT) layers are inserted into the model to embed edge information properly. The proposed model consists of an edge extractor (EdgeNet), EC-CNN blocks, and a DeepLabV3 \cite{chen2017rethinking} based semantic segmentation network. As an edge extractor, HED (Holistically-nested Edge Detection) \cite{xie2015holistically} is employed to obtain high-quality edge information. However, there is no thermal dataset with edge annotations; the RGB dataset was used for HED training. Even though HED is trained on an RGB dataset with ground truth edge annotations, the edge results of thermal images are quite successful. EC-CNN blocks consist of convolutional layers and GFT layers to guide the segmentation of the input image by using the output of the EdgeNet. Also, the DeepLabV3 model employs ResNet as feature extractor and atrous convolutions, whereas some ResNet blocks are replaced with EC-CNN block to embed edge prior.

Wang et al. \cite{wang2019thermal} proposed a thermal infrared pedestrian segmentation algorithm including a conditional generative adversarial network (IPS-cGAN). The generator of the IPS-cGAN is based on Unet \cite{ronneberger2015u} with two modifications so that a more suitable network for thermal infrared pedestrian segmentation is obtained. Firstly, to have more efficient connections, original convolutional blocks are replaced by residual blocks. Secondly, 0.5 rate dropout has been deployed so that the network becomes more robust. Moreover, SandwichNet is designed with a symmetrical structure as the discriminator of the proposed network. SandwichNet takes original image and segmentation results as inputs. The SandwichNet is designed based on multi-channel input PatchGAN \cite{isola2017image}. The difference is that SandwichNet needs symmetrical three-channel result-image-result with segmentation result from the generator and thermal image, and three-channel truth-image-truth with segmentation ground truth. Moreover, the designed generator and the discriminator are trained as an end-to-end GAN algorithm with cross-entropy loss. The modifications on Unet and the design of the discriminator provide a more robust model against noises for thermal infrared pedestrian segmentation. 

The combination of the Gaussian-Bernoulli Restricted Boltzmann Machine (GB-RBM) and convolutional neural network is proposed in RT-SegRBM-Net \cite{masouleh2019development} to segment the vehicles from the UAV-based thermal images in real-time. The deep learning algorithm is designed based on SegNet \cite{badrinarayanan2017segnet} architecture, and GB-RBM is embedded into the overall structure to make use of the geometry information of the vehicles.

Nightvision-Net (NvNet) \cite{chen2020nv} is proposed for semantic segmentation of low-resolution infrared images in weak illumination environmental conditions. Nv-Net suggests the network architecture of the FCN-8S \cite{bhandari2019context} with a contracting and an expanding path and a weighting loss. Also, transfer learning is utilized to increase the performance of semantic segmentation. NvNet architecture consists of four parts, data refinement (DR), data normalization (DN), the contracting path, and the expanding path. The contracting path has several convolution layers and average pooling operations and outputs down-sampled feature maps. Therefore, the aim of the expanding path is to enhance the output feature map's resolution. The contracting path is complemented by the expanding path that applies consecutive layers with upsampling operations instead of pooling operations. Also, the expanding path uses the feature maps from the corresponding layers of the contracting path to achieve better localization of the objects. Moreover, data normalization is performed, which accelerates the convergence of the training. Besides, NvNet introduces weighted-sigmoid-cross-entropy loss to calculate the error between the prediction and ground truth.

Xiong et al. proposed a Multi-level correction network (MCNet) \cite{xiong2021mcnet} to achieve thermal images segmentation for nighttime driving scenes. Thermal images have low resolution and blurred edges caused by the thermal crossover; therefore, MCNet proposes the multi-level attention module (MAM) to solve this problem. The MAM includes two sub-modules, the context aggregation module (CAM) and the correlation matrix correction module (CMCM). CAM is chosen to model the spatial correlations within pixels' position, and the correlation matrix learns the dependency between any two pixels. The correlation matrix has significant importance since the properties of the thermal images, such as low resolution and ambiguous object boundaries, may cause misleading results about the related contextual information. So, to prevent this misleading information and suppress the noisy information, the CMCM module is also included in the proposed method. If the correlation values between the intra-class pixels are lower than that of inter-class pixels, the CMCM module corrects these wrong values. Furthermore, thermal images are more dependent on edge information due to the lack of color information. Hence, a multi-level edge enhancement module (MEEM) is designed to enhance the edge information and improve the final feature representation in multiple iterations.

Feature Transverse Network (FTNet) \cite{panetta2021ftnet} is an end-to-end trainable convolutional neural network architecture. FTNet employs an encoder-decoder structure and an edge guidance part to conduct reliable pixel-wise classification. FTNet introduces a feature transverse network (decoder) exploiting a set of residual units \cite{he2016deep}. Moreover, ResNeXt \cite{xie2017aggregated} based encoder network provides thermal image features at different resolutions by subsampling at several stages. These feature maps are passed through the aforementioned residual units. Also, FTNet employs a fully connected layer to combine the outputs of the residual units. Edge information is also exploited to reduce the effects of thermal crossover and noise created by the sensors on the segmentation map. Moreover, the edges are extracted from the feature map obtained in the third layer of the encoder and passed through the combination of layers convolution, batch normalization, ReLU, respectively. Then, the edge map is upsampled to the input image resolution before passing through another convolutional layer. Finally, the edge map is also fused with the feature maps obtained in the decoder, and the result is passed through the final block, including convolutional, batch normalization, and ReLU layers. In addition, an edge-based loss function is adapted with the semantic loss while training FTNet to increase the segmentation accuracy. Edge ground truths are calculated from the semantic label gradients. 

\subsection{Analysis \& Results}

Thermal images alone can provide sufficient information in adverse environmental conditions, so a few segmentation methods have been developed using only thermal images. Although thermal crossover and noise introduced by the thermal imaging sensors make the segmentation task more difficult, the thermal segmentation methods propose different approaches such as employing edge information and correlation matrix. \cite{li2020segmenting}, \cite{xiong2021mcnet} and \cite{panetta2021ftnet} propose mechanisms to extract the edge information from the thermal image and use it to guide segmentation. \cite{masouleh2019development}, \cite{chen2020nv}, and \cite{wang2019thermal} employ encoder-decoder structures, whereas \cite{li2020segmenting} exploits atrous convolutions to obtain the output segmentation map. Moreover, \cite{xiong2021mcnet} creates the correlation matrix, which models the dependency between any two pixels, to focus on the same classes and avoid noisy information. Also, \cite{chen2020nv} exploits weighted-sigmoid-cross-entropy loss for images collected in weak illumination environmental conditions to discriminate important pixels while calculating the loss. \cite{masouleh2019development} attempts to segment vehicles from UAV-based thermal images, so a Boltzmann machine is employed for geometry extraction from vehicles up view to increase the segmentation accuracy. \cite{xiong2021mcnet} is proposed for nighttime driving scenes, so the model exploits inherent aspects of driving scene images, such as the fact that object instances show only in narrow bands that cross horizontally through the image's center. 

SODA dataset \cite{li2020segmenting} includes day and night thermal images for generic purposes and commonly used for testing thermal segmentation methods. On the SODA testing set, \cite{li2020segmenting} reported the performance of the proposed method and \cite{chen2017rethinking} as 0.619 and 0.571 mIoU, respectively. Moreover, \cite{xiong2021mcnet} and \cite{panetta2021ftnet} reaches 0.503 and 0.600 mIoU on the same dataset, as reported in \cite{panetta2021ftnet}. It can be noted that \cite{li2020segmenting} and \cite{panetta2021ftnet} have comparable results on SODA. In addition, \cite{wang2019thermal} is designed to overcome regional intensity inhomogeneity and be more robust against various noises for the infrared pedestrian segmentation. According to the revealed results in \cite{wang2019thermal}, the proposed method performs with 0.939 mIoU on its own dataset, and outperforms \cite{chen2017rethinking}, \cite{mirza2014conditional} and \cite{isola2017image}. In terms of the average precision results, \cite{masouleh2019development}, \cite{he2017mask}, \cite{chen2017rethinking}, and \cite{badrinarayanan2017segnet} achieves the similar performance on the NPU\_CS\_UAV\_IR\_DATA \cite{liu2018real} test sets, whereas \cite{masouleh2019development} achieves slightly better average processing time, as reported in the \cite{masouleh2019development}. In addition, \cite{chen2020nv} reported that the proposed model performs with 0.912 mIoU, which is better than \cite{chen2017deeplab} with 0.469 mIoU on the LII dataset. On the SCUT-Seg nighttime driving dataset, \cite{xiong2021mcnet} reported 0.676 mIoU and 32.52 FPS with a single NVIDIA GTX 1080 Ti. Moreover, \cite{panetta2021ftnet} announces its accuracy as 0.667 mIoU on the same dataset. Similar to SCUT-Seg, MFNet dataset \cite{ha2017mfnet} also contains driving scenes, and \cite{xiong2021mcnet} reaches 0.519 mIoU on the thermal images in MFNet dataset, while \cite{chen2018encoder} only achieves 0.504 mIoU and RTFNet-50 \cite{sun2019rtfnet} using both RGB and thermal images achieves 0.503 mIoU according to the revealed results in \cite{xiong2021mcnet}. Using the thermal images in the \cite{ha2017mfnet} dataset, \cite{panetta2021ftnet} reported its accuracy as 0.471 mIoU which is also comparable with the results of \cite{xiong2021mcnet} and \cite{sun2019rtfnet}. 

\section{Conclusion}
This survey reviews recent progress in deep learning-based semantic segmentation methods using thermal images and compares them in terms of their architectures, performance, applications, and the proposed approaches to improve the models. Also, this survey provides thermal image datasets descriptions.

In conclusion, using thermal images in semantic segmentation tasks helps to increase the robustness and success of the systems. Also, the proposed methods can be used in a wide range of applications. Due to the limited number of available thermal image datasets and characteristics of the images, only a few methods have been developed. The semantic segmentation of thermal images is very promising, and further research can be advanced in several directions, such as creating synthetic data, data augmentation, and fusion strategies. 

{\small
\bibliographystyle{ieee_fullname}
\bibliography{egbib}
}

\end{document}